\documentclass{article}

\usepackage{arxiv}

\usepackage[utf8]{inputenc} 
\usepackage[T1]{fontenc}    
\usepackage{hyperref}       
\usepackage{url}            
\usepackage{booktabs}       
\usepackage{amsfonts}       
\usepackage{nicefrac}       
\usepackage{microtype}      
\usepackage{graphicx}	

\title{A Dynamic Reduction Network for Point Clouds}

\author{
  Lindsey Gray, Thomas Klijnsma\\
  Fermi National Accelerator Laboratory\\
  Batavia, IL 60510 \\
  \texttt{lagray@fnal.gov} \\
  \And
  Shamik Ghosh \\
  Saha Institute of Nuclear Physics\\
  Kolkata, West Bengal, India \\
  \texttt{shamik.ghosh@cern.ch} \\
}

\begin{document}
\maketitle

\begin{abstract}
Classifying whole images is a classic problem in machine learning, and graph neural networks are a powerful methodology to learn highly irregular geometries.
It is often the case that certain parts of a point cloud are more important than others when determining overall classification.
On graph structures this started by pooling information at the end of convolutional filters, and has evolved to a variety of staged pooling techniques on static graphs.
In this paper, a dynamic graph formulation of pooling is introduced that removes the need for predetermined graph structure.
It achieves this by dynamically learning the most important relationships between data via an intermediate clustering.
The network architecture yields interesting results considering representation size and efficiency. 
It also adapts easily to a large number of tasks from image classification to energy regression in high energy particle physics.\footnote{This preprint will be updated with more studies on different open datasets.}\footnote{The code needed to reproduce the results in this text is included in the arxiv submission file.}
\end{abstract}


\section{Introduction}

Pooling has been an essential component of modern machine learning, allowing pertinent local information to be propagated to global intermediate feature sets or final discriminators.
The shape of the pooling operation is typically determined by hand, setting the size of a convolutional filter and the number of pooling steps before an output layer.
This process is difficult to optimize for graph neural networks~\cite{Bronstein_2017}, since neighbourhoods of nodes may vary in size and meaning depending on the problem at hand.
In the area of message passing neural networks~\cite{gilmer2017neural} there are recent advancements in learned pooling techniques on graphs~\cite{diehl2019edge}, and there is small but steady progress on using pooling to alter input graph structures.
This text describes a new pooling architecture using dynamic graph convolutions~\cite{wang2018dynamic} and clustering algorithms to learn an optimized representation and corresponding graph for pooling.
The model used in the following text is implemented in Pytorch Geometric~\cite{Fey/Lenssen/2019} (PyG).

This architecture was derived in the context of hadron\footnote{Particles that are bound together by the strong force.} energy regression in High Energy Physics (HEP), where graph neural networks are beginning to solve difficult clustering problems~\cite{gravnet, NEURIPS2019} in novel ways.
The objective of that problem is to determine the original energy of a particle incident upon a device called a ``sampling calorimeter." 
Within the calorimeter, which is made of dense material like lead or steel interspersed with lighter material, incident particles above a threshold energy will produce pairs of particles by nuclear interaction, creating a ``shower" of particles.
At a number of fixed depths within the calorimeter, it records scintillation or ionization signals at fixed depths as a proxy for the number of produced particles.
These estimates of multiplicity can be used to infer the originating particle's energy.
The energy deposition patterns of hadrons are known to have a high degree of local fluctuations in particle multiplicity during the shower's evolution.
This means that throughout the shower there are randomly located regions that require different treatment from more homogeneous ones, and so there exists an optimal dynamic cluster for each hadron shower's data to best estimate the energy.

A human-designed algorithm called ``software compensation"~\cite{Tran_2017} has been developed to solve this problem as well.
It is already based in the principle of reducing an objective function to generate learned weights to determine shower energies.
However, there is a significant amount of specific tuning that needs to be done to make the algorithm function for varying detector designs, and its domain of applicability remains well within HEP.
Using the technique described in this paper, the entire algorithm is now learned, rather than a specific part of a correction, and the algorithm can dynamically adapt to the topology of a given hadron shower.
Using a machine learning algorithm for this task mitigates the need for manual specialization, and affords the possibility to investigate the applications of technique on rather different tasks from calorimetry and with more widely available datasets.
Benchmarks for various estimation and classification tasks will be demonstrated in the more classic machine learning tasks.

The advantages of this dynamic reduction architecture are:
\begin{itemize}
    \item Representation spaces with good performance are very small.
    \item No prior graph structure is necessary, and if one is provided it can be altered by the pooling layers, since the graph pooling structure is learned.
    \item Without a prior graph structure, data for training and inference need very little preprocessing beyond normalization and stacking.
\end{itemize}

\section{Related Work}

The architecture proposed here is similar in outcome to the techniques proposed in~\cite{diehl2019edge, lee2019selfattention}, but radically different in implementation.
Our focus is on learning latent representations that optimize the pooling performance of an unsupervised clustering algorithm.
In particular, previous works in graph learning~\cite{monti2016geometric,fey2017splinecnn} demonstrate that controlling the behavior of an unsupervised algorithm can help in learning concise representations quickly. 
However, aspects of the original structure of the data were kept and only messages to pass in that structure were generated.
This text expands on both previous aspects by combining it with dynamic graph convolutions~\cite{wang2018dynamic} and controlling a clustering algorithm in the latent space to produce a dynamically learned optimized pooling.

\section{Dynamic Reduction Network}

\begin{figure}[!hbt]
    \centering
    \includegraphics[width=0.8\textwidth]{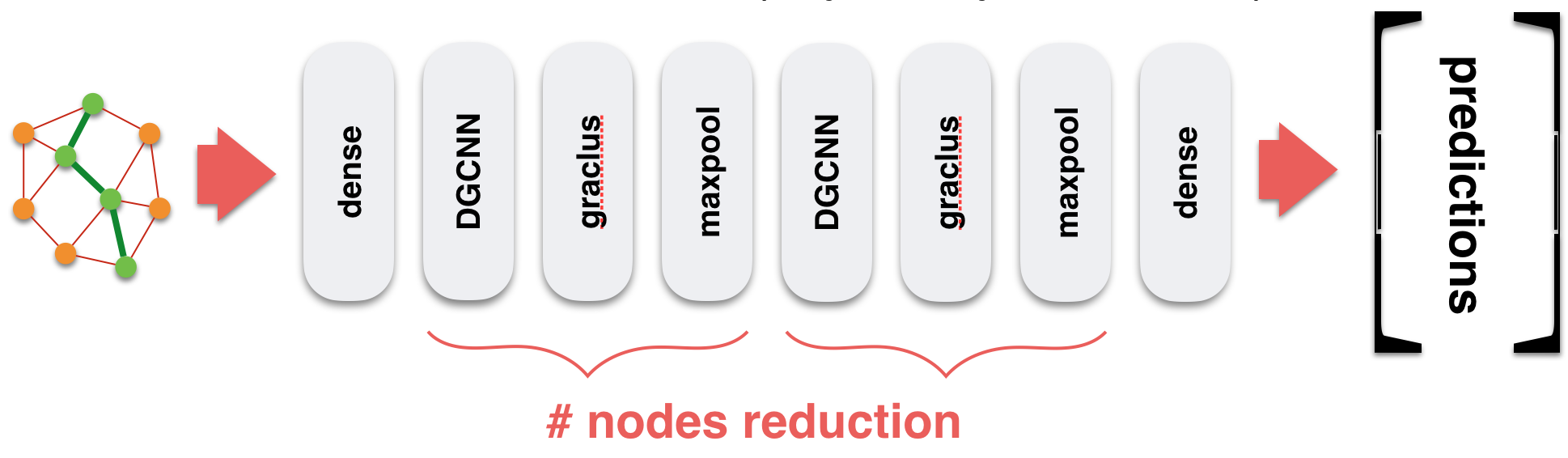}
    \caption{The basic workflow of the Dynamic Reduction Network, details given below.}
    \label{fig:drn_flow}
\end{figure}

All clustering algorithms can be treated as an indexing of nodes, so any clustering algorithm can be used in the latent space to make this demonstration.
The unsupervised clustering algorithm is treated as a black box, and the supervised task is to optimize its performance for the task at hand.
Using a dynamic graph convolutional approach,  it is possible to learn the parameters that maximize the impact of the clustering on reducing information for further processing, and hence this network is called a dynamic reduction network (DRN).
We have chosen to use a readily available greedy popularity based clustering algorithm~\cite{10.5555/2340646.2340660} as an initial demonstrator.
Other clustering algorithms will be tested and compared once their GPU implementations are made available in PyG in the future.

The default model used for the MNIST superpixels~\cite{monti2016geometric} classification task in this paper is composed as follows:
\begin{enumerate}
    \item The input data are normalized by fixed values such that the input data largely occupy the range $[0,1]$, outliers are allowed.
    \item A multilayer perceptron (MLP) encoding the normalized input data to a latent space of dimension N is applied to all input nodes. The default depth is 3 layers, with an intermediate layer half the size of the final output.
    \item The latent space data are processed by a dynamic graph convolution layer, i.e. neighbours are found in the latent space rather than the original representation. The internal messages are created using a MLP with three layers, starting from 2N, 1.5N, and outputing a message of width N. The update function can be summation, maximum, or average.
    \item The resulting latent nearest neighbours graph is then weighted by distance and clustered using a greedy clustering algorithm, pooling the node features by taking the maximum of the clustered data.
    \item The reduced latent data are passed through another dynamic convolutional filter, and clustered again with the same algorithm.
    \item The results of the second learned pooling step are then globally max pooled and passed through an MLP decoder to produce the output logits.
\end{enumerate}
\noindent
This process is summarized in Figure~\ref{fig:drn_flow}.
Alterations to this model used for various test are described later in the text.
The depth of the MLPs in the various encoding, message passing, and decoding steps are parameterized.
The number of nearest neighbours, k, is a hyper-parameter of the model and may require tuning to a given task, depending on the relational data that is needed to make a prediction.

\section{Results}

This model was tested on MNIST ``superpixels" dataset with 75 superpixels~\footnote{This model was developed using private data of the CMS Collaboration which cannot be published here.}.
The superpixels dataset is a downsampled and aggregated form of the full MNIST dataset.
Previous graph models are shown in Refs.~\cite{monti2016geometric, fey2017splinecnn}, 
A result of a scan in hidden-dimension (the ``width") of the network is shown in ~\ref{fig:MNISTSP_perf}. 
During the training and evaluation both the original graph structure and all pixels that are not filled are dropped, the data are ``zero-suppressed".
Each superpixel has a pair of coordinates defining a centroid, and an intensity.
The performance using a width of 20 channels with $\mathrm{k} = 4$ is a factor of two better on 75 superpixels than the reference models in ~\cite{monti2016geometric, fey2017splinecnn}, and approaches the performance of models trained on full MNIST at a width of 256 channels.
The data passed to the model consists only of the centroid x and y, and the gray-scale of that pixel. 
In Figure~\ref{fig:MNISTSP_perf} each DRN is trained for 400 epochs using an nVidia Tesla V100 with the AdamW optimizer~\cite{loshchilov2017decoupled} using one-cycle cosine annealing (on the learning rate only) with a starting value of 0.001.
The weight decay is held constant at a value of 0.001.
Best performance is often achieved quite early, so further tuning of model training and optimization is possible, and volatile GPU usage is low.
There is significant room for improving the training and evaluation time performance of this model.

\begin{figure}[!hbt]
    \centering
    \begin{tabular}{c|c|c|c|c}
        Model & No. of & Best achieved & Epoch of best & Performance\\
        variant & parameters & performance & performance & at 400 epochs \\ \hline
        DRN20 & 5123 &  \emph{0.9761} & 180 & 0.9731 \\
        DRN32 & 12797 & 0.9806 & 205 & 0.9792 \\
        DRN64 & 50157 & 0.9872 & 347 & 0.9866 \\ 
        DRN128 & 198605 & 0.9891 & 232 & 0.9884 \\
        DRN256 & 790413 & $\mathbf{0.9905}$ & 217 & 0.9899 \\
        MoNet & -- & 0.9111 & -- & -- \\
        SplineCNN & 63786 & 0.9522 & 40 & -- \\
    \end{tabular}
    \caption{Performance for a scan in dynamic reduction network width of the architecture described above on the MNIST ``superpixels" dataset~\cite{monti2016geometric} with 75 superpixels, and two established models. DRN<N> means a dynamic reduction network with hidden dimension N is used. Comparison to similar models on the superpixels dataset are made. All superpixel graph data has been dropped and the network is allowed to learn an entirely new representation based on the pixel data. For all DRN models, k = 4. The performance at a hidden dimension size of 20 is particularly interesting given the modest gains in performance from wider networks. `--' indicates unknown data.}
    \label{fig:MNISTSP_perf}
\end{figure}

The superpixels dataset is known to yield poor performance in CNN-based networks, and graph networks have been proposed in previous work to improve upon this. 
The performance demonstrated in Figure~\ref{fig:MNISTSP_perf} shows clearly that further improvements on very under-sampled data are achievable and this model sets new performance benchmarks on these datasets, especially in terms of model size.

\section{Conclusions}

This text introduces the Dynamic Reduction Network as a new tool in High Energy Particle physics for processing sampled data from a highly varying multi-dimensional image.
This is accomplished by designing a network that can learn effective pooling strategies by manipulating an unsupervised algorithm in a high-dimensional latent space.
To demonstrate the efficacy of this network, a performance benchmark on an undersampled MNIST dataset indicates that this new architecture outperforms previous graph based architectures, even for very small numbers of parameters.
This outcome suggests a powerful new technique for approaching classification and regression problems in both computer vision and high energy physics.

\section{Acknowledgements}

This research was supported in part by the Office of Science, Office of
High Energy Physics, of the US Department of Energy under Contract No. DE-AC02-07CH11359 through FNAL LDRD-2019-017.

Many thanks to Matthias Fey and Song Han for quick consultation on this model.
Thanks as well to Nhan Tran and Salvatore Rappoccio for proofreading assistance.

\bibliographystyle{unsrt}  
\bibliography{references} 

\end{document}